\documentclass[conference]{IEEEtran}
\IEEEoverridecommandlockouts
\usepackage{cite}
\usepackage{amsmath,amssymb,amsfonts}
\usepackage{algorithmic}
\usepackage{graphicx}
\usepackage{textcomp}
\usepackage{xcolor}
\usepackage{booktabs}
\usepackage{amsmath}
\usepackage{multirow}
\usepackage{float}
\usepackage{tabularx}
\usepackage{dblfloatfix}
\usepackage{placeins}
\usepackage{tabularx}

\def\BibTeX{{\rm B\kern-.05em{\sc i\kern-.025em b}\kern-.08em
    T\kern-.1667em\lower.7ex\hbox{E}\kern-.125emX}}
\begin{document}

\title{A self-attention based deep learning method for lesion attribute detection from CT reports}

\author{\IEEEauthorblockN{Yifan Peng\IEEEauthorrefmark{1}, 
Ke Yan\IEEEauthorrefmark{2}, 
Veit Sandfort\IEEEauthorrefmark{2}, 
Ronald M. Summers\IEEEauthorrefmark{2}, 
Zhiyong Lu\IEEEauthorrefmark{1}}
\IEEEauthorblockA{\IEEEauthorrefmark{1}\textit{National Center for Biotechnology Information, National Library of Medicine, National Institutes of Health}, Bethesda, USA \\
\IEEEauthorrefmark{2}\textit{Department of Radiology and Imaging Sciences, Clinical Center, National Institutes of Health}, Bethesda, USA\\
\{yifan.peng, ke.yan, veit.sandfort, rms, zhiyong.lu\}@nih.gov}
}

\maketitle

\begin{abstract}
In radiology, radiologists not only detect lesions from the medical image, but also describe them with various attributes such as their type, location, size, shape, and intensity. While these lesion attributes are rich and useful in many downstream clinical applications, how to extract them from the radiology reports is less studied. This paper outlines a novel deep learning method to automatically extract attributes of lesions of interest from the clinical text. Different from classical CNN models, we integrated the multi-head self-attention mechanism to handle the long-distance information in the sentence, and to jointly correlate different portions of sentence representation subspaces in parallel. Evaluation on an in-house corpus demonstrates that our method can achieve high performance with 0.848 in precision, 0.788 in recall, and 0.815 in F-score. The new method and constructed corpus will enable us to build automatic systems with a higher-level understanding of the radiological world.
\end{abstract}


\section{Introduction}
\begin{figure}
    \centering
    \includegraphics[width=.48\textwidth]{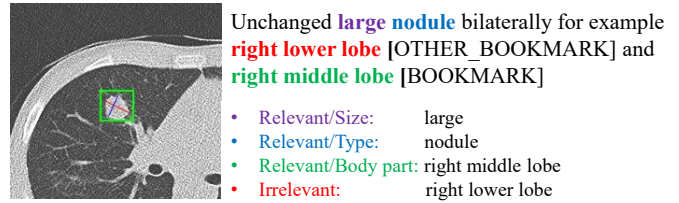}
    \caption{Sample sentence with bookmarks.}
    \label{fig:sample}
\end{figure}

In radiology, finding lesions in the imaging study and describing them in the radiology report is the main task for the radiologists. The description usually contains rich attributes such as associated body part, type, and size. Take Fig.~\ref{fig:sample} as an example. While interpreting a CT scan (on the left), the radiologist describes the lesion with the sentence ``Unchanged large nodule $\dotsc$ right middle lobe \textbf{\underline{BOOKMARK}}'' and places a hyperlink (hereafter ``bookmark'') in the context to refer to the specified lesion in the image. Here, ``right middle lobe'' is the body part, ``nodule'' is the lesion type, and “large” is the size attribute.

The attributes of lesions in the radiology report are informative and useful in various tasks. In image-based computer-aided diagnosis, we can use the attributes to train fine-grained lesion image classification models. The content-based lesion retrieval, on the other hand, can use them to retrieve similar lesion images when their appearances in the CT scans are not strictly identical. In clinical NLP, the lesion with attributes can help build the semantic graph of each report to better interpret how the radiologist reached the impressions from findings.

Although the lesion-associated semantic attributes are important, annotating them in the radiology reports is time-consuming and expensive. Besides, extracting these attributes from the sentences is non-trivial. First, the sentences often contain a complex mixture of information describing not only the bookmarked lesion of interest but also other related lesions (hereafter ``other bookmarks''). For example, in Fig.~\ref{fig:sample}, there are 4 labels matched based on the ontology, namely ``large'', ``nodule'', ``right lower lobe'', and ``right middle lobe''. Among them, the label ``right lower lobe'' is irrelevant since it describes another lesion. Second, there are also uncertain labels in the sentences, such as ``adenopathy or mass''.

In this paper, we formally call a label ``relevant'' if it describes the bookmark of interest, ``irrelevant'' if it describes other bookmarks, and ``uncertain'' if it is in a hypothetical statement. Since both the irrelevant and uncertain labels may bring the noise to downstream training, it is important to distinguish them from relevant labels.

To tackle these challenges, this paper outlines a new text-mining method to automatically extract relevant bookmarked-specified attributes in the sentence. That is, given a sentence with multiple attributes and bookmarks, we aim to assign relevant labels to each bookmark from all label-bookmark pairs. Consequently, we reformulate this task as a relation classification problem and propose to use a self-attention based deep neural network because of its superior performance in various NLP tasks in recent years. 

We evaluate the performance of the proposed method on an in-house corpus with 1,890 sentences manually annotated by two expert radiologists. Our method obtained 0.848 in precision and 0.788 in recall for an F-score of 0.815, demonstrating the effectiveness of machine learning-based approaches for automatic relation extraction from the clinical text in this task.

\section{Related work}
\begin{figure*}
    \centering
    \includegraphics[width=\textwidth]{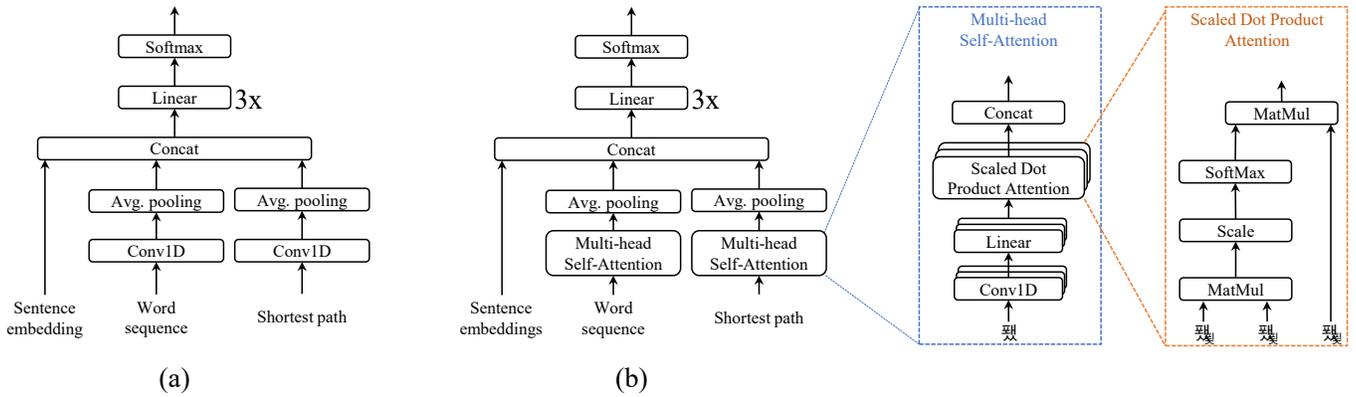}
    \caption{Architectures of the deep learning models. (a) CNN. (B) Multi-head CNN.}
    \label{fig:archetectures}
\end{figure*}
In recent years, there has been considerable interest in examining the rich clinical information stored in electronic health records. However, manually annotating a large dataset to fulfill the data-hungry deep learning models is time-consuming and expensive. For example, a radiologist usually read at a speed of 10 mins per example for CT scan studies~\cite{sokolovskaya2015effect}. Instead, researchers may benefit from using text-mining to generate annotations even if those annotations are of modest accuracy~\cite{wang2017chestx}. To reduce manual annotation burden, some researchers leveraged the rich information contained in associated radiology reports. Disease-related labels have been mined from reports for classification and weakly-supervised localization on X-ray and CT images~\cite{tang2018attention,li2003learning,shin2016interleaved,yim2017classifying}.

Owing to the rapid growth of available EHR, deep learning methods for text mining become more appealing recently because of its competitive performance versus traditional methods and its ability to relieve the feature sparsity and engineering issue~\cite{shickel2018deep}. For example, both multi-channel dependency-based CNNs~\cite{peng2017deep} and shortest path-based CNNs~\cite{hua2016shortest} are well suited for sentence-based relation extraction. It is also generally faster to train a CNN model than other deep learning networks.

Despite its efficiency, the main limitation of CNN is its shortcoming to capture long-distance information. This is due to the fixed window size in the convolutional layers. Several studies have been conducted to solve this problem by introducing linguistic information (e.g., shortest path) in the input layer or multichannel to capture the hierarchy structure of the sentence~\cite{peng2018extracting}.

More recent attempts have been made to utilize the attention mechanism to help the network focus on salient features~\cite{xu2015show}. It has been firstly used in recurrent neural networks architectures in the NLP applications like machine translations. In those cases, the attention mechanisms allow the model to ``attend to'' (correlate) different parts of the sentence at each step, thus the output depends on a weighted combination of all the input states. Vaswani et al. extended this idea by proposing the Transformer, a model that relies on a self-attention mechanism to draw global dependencies between input and output~\cite{vaswani2017attention}. They also introduced ``multi-head'' to attend different portions of the representation subspaces in parallel. Gao et al. further demonstrated that the self-attention mechanism can be used in the CNN-based approaches, achieving both fast and accurate performance in the text classification task~\cite{gao2018hierarchical}.

In this work, we hypothesize that a similar self-attention mechanism could be used in the clinical relation extraction tasks. In the following sections, we show how we adapt it to our problem.

\section{Method}

This task focuses on distinguishing bookmark-relevant labels from irrelevant and uncertain ones from the sentences. To tackle this problem, we convert it to a relation classification problem. Given a sentence with multiple labels and bookmarks, we construct all label-bookmark pairs as candidates. For example, the pair of ``BOOKMARK, right middle lobe'' is relevant, but the pair of ``BOOKMARK, right lower lobe'' is irrelevant. The goal of this task is thus to predict whether the pairs are relevant, irrelevant, or uncertain.

We propose to address this task using a CNN model which has been widely used in the relation extraction task (Fig.~\ref{fig:archetectures}). The input of our model consists of two parts, the word sequence with the mentioned attribute and BOOKMARK, and the sentence embedding. The model outputs a probability vector (three elements) corresponding to the type of the relation between the label and the bookmark (irrelevant, uncertain, and relevant). Our model consists of three layers: a word embedding layer, a convolution layer, and three fully-connected layers. 

\subsection{Embedding layer}

\subsubsection{Word sequence}

The first input is the word sequence. Each word in a sentence is represented by concatenating its word embedding, part-of-speech, chunk, named entity, and position features. Here we use three attributes as named entities: size, type, and body part. The part-of-speech, chunk, and named entity features are encoded using a one-hot schema. We also used the position feature proposed in~\cite{sahu2016relation}, which consists of two relative distances, $d_1$ and $d_2$, for representing the distances of the current word to the attribute and BOOKMARK respectively. Both $d_1$ and $d_2$ are non-linearly mapped to a ten-bit binary vector, where the first bit stands for the sign and the remaining bits for the distance~\cite{zhao2016drug}.

\subsubsection{Shortest-path}

Besides the word sequence, we also used the shortest dependency path between BOOKMARK and attributes as the input. The representation of each word along the path is the same as in the word sequence except the position features which are then calculated based on the shortest path instead. The shortest-path is widely used to extract the long-distance dependencies in the sentence~\cite{bunescu2005shortest}. Such information is considered overlooked by classical CNN due to the small window size.

\subsubsection{Sentence embedding}
We also used sentence embeddings to capture the sentence semantics. Sentence embeddings have shown promising results recently as the semantic is represented by high dimensional vectors. Such vector-based representations are commonly learned from large text corpora and have become increasingly important in recent clinical text-mining studies~\cite{chiu2016how,chen2018biosentvec}.

\subsection{Multi-head self-attention}

Following the work of~\cite{vaswani2017attention,gao2018hierarchical}, we used multi-head self-attention to discover the relations between entries in the sequence. 
\begin{align}
    \text{Multi-head}(E) & = [\text{head}_1, \dots, \text{head}_h]\\
    \text{head}_i & = \text{softmax}(\frac{E_iE^T_i}{\sqrt{d_i}})E_i\\
    E_i &= \text{ELU}(\text{Conv1D} (E_i ) W_i^q+b^q)
\end{align}
where $E \in \mathbb{R}^{l\times d}$ is the input from the embedding layer.

The method first splits the embedding input into h parts, each of which attends to a different portion of the embedding dimension. Multi-head attention allows the model to attend to information from different portions of the embeddings so that the final output sequence can be constructed from a more expressive combination.

For each head, the method then uses a scaled-dot-product attention to discover the relations between words in the sequence~\cite{vaswani2017attention}. The intuition behind this is to first calculate a weight matrix $E_i E_i^T$  based on the similarity of words in the sequence, then multiply back with $E_i$ so that each word is a weighted average of all words in the input sequence. By doing so, the model draws global dependencies between words. 

Rather than use the embedding input directly, we applied convolution with $d_i$ filter maps and a window size of 3 to extract features from the embedding inputs to get local features $E_i\in\mathbb{R}^{l\times d_i}$, where $l$ is the length of the sequence and $d_i$ is the dimensionality of the embeddings' $i$th heading. This will provide a piece of context information for each word in the sequence.

Like in~\cite{gao2018hierarchical}, we applied the exponential linear units (ELUs) as the activation function. The reason is that ELUs can output negative values that allow a larger range of values compared to Rectified Linear Unites (ReLUs).

\subsection{Sentence hierarchy}

Both sentence features and shortest-path features then feed into a global average pooling layer across the entire sentence. In this task, we find that using average pooling outperforms using max-pooling. Afterward, the sentence features and shortest-path features were combined with the sentence embeddings, followed by three fully-connected layers.

\section{Experiments}

\subsection{Dataset}

In this section, we described the process to construct gold-standard labels from the radiology report associated with lesions on CT images. 

First, we constructed an attribute list of interest based on RadLex~\cite{langlotz2006radlex}. The list contains frequently mentioned attributes mentioned in the radiology reports such as body part (e.g., chest, abdomen), types (e.g., nodule), and size, shape and intensity of the lesions. The list was then verified by the radiologists listed as authorship. The final vocabulary consists of 171 attributes.

After constructing the lesion attribute vocabulary, we randomly selected 1,890 sentences with at least one bookmark from the DeepLesion dataset~\cite{yan2018deeplesion} and extracted all attribute mentions based on the vocabulary. Specifically, we first tokenized the sentence and lemmatized the words in the sentence using NLTK to obtain their base forms~\cite{bird2009natural}. We then matched the attribute mentions in the preprocessed sentences and normalized them.

Finally, we created all ``bookmark, attribute'' pairs as candidates and asked a physician (VS) to annotate their relevance (relevant/uncertain/irrelevant). As a result, we obtained the gold standard corpus with 7,356 relevant, 477 uncertain, and 1,464 irrelevant. We then use 60\% for training, 20\% for development, and 20\% for testing (Table~\ref{tab:statistics}).
\begin{table}[!t]
    \caption{Statistics of the corpus\label{tab:statistics}}
    \begin{center}
    \begin{tabular}{lrrr}
\toprule
 & \textbf{Training} & \textbf{Validation} & \textbf{Test}\\
\midrule
Sentences & 1,144 & 370 & 376\\
Instances &  &  & \\
\hspace{2em}Relevant & 4,418 & 1,448 & 1,490\\
\hspace{2em}Uncertain & 326 & 48 & 103\\
\hspace{2em}Irrelevant & 868 & 291 & 305\\
\bottomrule
    \end{tabular}
    \end{center}
\end{table}
\begin{table}
    \caption{Regular expressions used in the rule-based system\label{tab:rules}}
    \begin{center}
    \begin{tabularx}{.48\textwidth}{l@{~}l@{~}X}
\toprule
Type &  & Regular Expression\\
\midrule
Irrelevant & \textbullet & (no evidence of $|$ no evidence of developing $|$ no evidence of abdominal $|$ not $|$ poorly $|$ previously seen $|$ without $|$ without evidence of) ATTRIBUTE\\
 & \textbullet & (adjacent to $|$ arising from $|$ above $|$ anterior to $|$ abutting $|$ beneath $|$ close to $|$ encasing $|$ left of $|$ left of this $|$ near $|$ posterior to $|$ right of) ATTRIBUTE\\
 & \textbullet & (other) ATTRIBUTE\\
Uncertainty & \textbullet & (or $|$ and / or $|$ / $|$ likely $|$ possibly) ATTRIBUTE\\
 & \textbullet & (dome of $|$ portion of $|$ tail of) ATTRIBUTE \\
\bottomrule
    \end{tabularx}
    \vspace{-2em}
    \end{center}
\end{table}
\begin{table*}[hbt!]
    \caption{Performance of the models. Performance is reported in terms of (P)recision, (R)ecall, AND (F)1-score.\label{tab:performance}}
    \begin{center}
    \begin{tabular}{l|rrr|rrr|rrr|rrr}
\toprule
 & \multicolumn{3}{c|}{Relevant} & \multicolumn{3}{c|}{Uncertain} & \multicolumn{3}{c|}{Irrelevant} & \multicolumn{3}{c}{Macro}\\
\cmidrule{2-13}
 & P & R & F & P & R & F & P & R & F & P & R & F\\
\midrule
Rule-based & 0.820            & 0.992 & 0.898 & 0.768            & 0.417 & 0.541 & 0.850      & 0.111       & 0.197 & 0.813           & 0.507 & 0.545\\
CNN & 0.951 & 0.899 & 0.924 & 0.523 & 0.544 & 0.533 & 0.673 & 0.843 & 0.748 & 0.716 & 0.762 & 0.735\\
CNN + rules & 0.954 & 0.899 & 0.926 & 0.534 & 0.602 & 0.566 & 0.675 & 0.839 & 0.749 & 0.721 & 0.780 & 0.747\\
Multi-head CNN & 0.934 & 0.971 & 0.952 & 0.718 & 0.544 & 0.619 & 0.863 & 0.764 & 0.810 & 0.838 & 0.760 & 0.794\\
Multi-head CNN + rules & 0.938 & 0.971 & 0.954 & 0.739 & 0.631 & 0.681 & 0.869 & 0.761 & 0.811 & 0.848 & 0.788 & 0.815\\
\bottomrule
    \end{tabular}  
    \end{center}
\end{table*}

\subsection{Rule-based system}

For comparison, we also implemented a rule-based system to detect irrelevant and uncertain labels (Table~\ref{tab:rules}). The rules were hand-crafted by heuristically investigating the validation set. Note that these rules do not use the information of bookmarks in the sentence.  When combined with the deep learning models, these rules are used for post-processing the output of the classifier.

\subsection{Experiment setup}

For our experiments, we used the Genia Tagger to obtain the part-of-speech, chunk tags, and named entities of each word~\cite{tsuruoka2005bidirectional}. We used pre-trained word embedding vectors and sentence embedding vectors learned on PubMed articles and MIMIC-III clinical notes using the fastText and sent2vec tools respectively~\cite{chen2018biosentvec}. We set the maximum sentence length to 128. That is, longer sentences were pruned, and shorter sentences were padded with zeros. For each fully-connected layer, we initialized the weights with Xavier normal initializer~\cite{glorot2010understanding}. We set the bias to be 0.01. To train the model, we used the Adam optimizer with a learning rate of 0.0007.  To prevent overfitting, we used dropout ($p = 0.5$). We also apply layer normalization after multi-head self-attentions~\cite{ba2016layer}. The model was run 10 epochs after the loss on validation set stop decreasing. For each epoch, we randomized the training examples and conducted a mini-batch training with a batch size of 128.

\subsection{Results and Discussion}

Table~\ref{tab:performance} shows the performance of the rule-based system, the CNN, the multi-head CNN, as measured by Precision, Recall, and F1-score. We also combined the CNN model and rule-based systems by applying rules for post-processing. 

Among these three models, the combination of Multi-head CNN and rules achieved the highest precision of 0.848, recall of 0.788 and F1-score of 0.815.

We observed that the results of the rule-based system and the deep learning method complement each other. The rule-based system tends to obtain high precision but lower recall (e.g., irrelevant attribute detection). On the other hand, the deep-learning method tends to be more balanced. Consequently, combining the two will dramatically improve the recall (up to 20\% over the rule-based system). Below are some long and complicated examples that our proposed model could correctly detect the relevant relation between the attribute and the bookmark.

\begin{itemize}
    \item There is a new right upper lobe \textbf{pulmonary nodule}, for example OTHER\_BOOKMARK and \textbf{\underline{BOOKMARK}}.
    \item Enlarged \textbf{mediastinal lymph node} remains stable in size, including a right paratracheal lymph node OTHER\_BOOKMARK, subcarinal lymph node conglomerate OTHER\_BOOKMARK, and right hilar lymph conglomerate \textbf{\underline{BOOKMARK}}.
\end{itemize}
	
We also performed an error analysis of the testing set. The most frequent errors (69.3\%) are because the classifier linked the attributes to the wrong bookmarks. The second most frequent errors (18.8\%) are a failure to capture uncertain attributes. Further analysis of these errors identified two main reasons. One is the parsing errors due to the complex structure of free-text radiology reports. For example in the sentence ``Smaller heterogeneously \textbf{enhancing} retrocrural nodule for example OTHER\_BMK left and \textbf{BOOKMARK}  right of the aorta'', the model failed to detect the relevant relation between ``enhancing'' and ``BOOKMARK'' because of the errors of parsing noun phrase conjunction ``OTHER\_BMK left and BOOKMARK right of the aorta''. The second reason is that the keyword is outside the scope of attribute and bookmarks. For example in the sentence ``There is a prevascular soft tissue nodular density which may represent a borderline enlarged \textbf{lymph node BOOKMARK}.'', the uncertain keyword ``may'' is way beyond the two entities.
%

\section{Conclusion}

In this paper, we studied a hybrid deep learning method to automatically detect attributes of lesions from the radiology reports. Evaluation on an in-house corpus demonstrated that our method can achieve high recall and precision. Future work includes the detection of more types of attributes, utilizing the keywords beyond the scope of interest, and evaluation of the method across corpora from multi-institutional radiology reports.

\section*{Acknowledgment}

This work was supported by the Intramural Research Programs of the NIH Clinical Center and National Library of Medicine. Research reported in this work was also supported by the National Library of Medicine of the National Institutes of  under award number K99LM013001-01. The content is solely the responsibility of the authors and does not necessarily represent the official views of the National Institutes of Health. We are also grateful to Dr. Shang Gao for sharing their codes.

\bibliographystyle{IEEEtran}
\bibliography{IEEEabrv,mybib}

\end{document}